\documentclass[11pt]{article}
\usepackage{lrec2006}
\usepackage{covington}
\usepackage{qtree}
\usepackage{rotating}

\title{A Corpus-based Study of Temporal Signals}
\name{Leon Derczynski and Robert Gaizauskas}

\address{University of Sheffield \\
                211 Portobello, S1 4DP, UK \\
                L.Derczynski@dcs.shef.ac.uk, R.Gaizauskas@dcs.shef.ac.uk}

\abstract{Automatic temporal ordering of events described in discourse has been of great interest in recent years. Event orderings are conveyed in text via various linguistic mechanisms including the use of expressions such as ``before", ``after" or ``during" that  explicitly assert a temporal relation -- {\it temporal signals}. In this paper, we investigate the role of temporal signals  in temporal relation extraction and provide a quantitative analysis of these expressions in the TimeBank annotated corpus.}

\begin{document}

\maketitleabstract

\section{Introduction}

The task of automatically determining the temporal relations that hold between events described in a text is a research challenge that has increasingly occupied researchers in computational language processing \cite{setzer2000annotating,pustejovsky2004specification,Ver09,verhagen2010semeval}. The mechanisms used to convey temporal relational information in text are complex and include tense, textual ordering, as well as specific lexical cues; and of course readers and writers bring to bear lexical and world knowledge, informing them of likely event sequences and inter-relationships.

Of the mechanisms that play a part in conveying temporal relational information, one that has been under-investigated is the use of expressions, typically adverbials or conjunctions, which  overtly signal temporal relations -- words or phrases such as {\it after}, {\it during} and {\it as soon as}. Very few of the teams participating in the recent TempEval challenges \cite{Ver09,verhagen2010semeval} exploited these words as features in their automated temporal relation classification systems.  Certainly no detailed study of these words and their potential contribution to the task of temporal relation detection has been carried out to date, despite their demonstrable utility~\cite{derczynski2010using}. This paper begins to address this deficiency. Using the TimeBank corpus, a corpus of news wire texts annotated with TimeML~\cite{pustejovsky2003timebank}, in which a class of expressions referred to as temporal signals is explicitly annotated, we set out to answer the following questions:
\begin{enumerate}
\item What proportion of temporal relations annotated in TimeBank have an associated temporal signal? That is, are explicitly signalled using a signal word or phrase?
\item Of the expressions which can function as temporal signals, what proportion of their usage in the TimeBank corpus is as a temporal signal? E.g. how ambiguous are these expressions in terms of  their role as temporal signals?
\item Of the occurrences of these expressions as temporal signals, how ambiguous are they with respect to the temporal relation they convey?
\end{enumerate}
The following paper provides provisional answers to these questions -- provisional as one of the difficulties we encountered was significant under-annotation of temporal signals in TimeBank. We have addressed this to some extent, but more work remains to be done. Nonetheless we believe the current study provides important insights into the behaviour of temporal signals and  how they may be exploited by computational systems carrying out the temporal relation detection task.

The remainder of the paper is divided into three parts. In section two we give a more detailed characterisation of temporal signals, further describe TimeBank and TimeML and discuss prior related work. In section three we describe the additional annotation work we have done on TimeBank and present the quantitative analysis that provides answers to the questions framed above. The fourth section considers, on a case by case basis, specific examples of expressions which are highly ambiguous as regards their role as temporal signals and discusses their behaviour in detail. 

\section{Temporal Signals}


%

\subsection{Linguistic Characterisation}
\label{signalstructure}

Signal expressions explicitly indicate the existence and nature of a temporal relation between two events or states or between an event or state and a time point or interval.  Hence a temporal signal has two arguments, which are the temporal "entities" that are related. One of these arguments may be deictic instead of directly attached to an event or time; anaphoric temporal references are also permitted. For example, the temporal function and arguments of \emph{after} in \emph{He slept \underline{after} a long day at work} are clear and available in the immediately surrounding text. With {\it After that, he swiftly finished his meal and left} we must look back to the antecedent of {\it that} to locate the second argument. 

Sometimes a signal will appear to be missing an argument; for example, sentence-initial signals with only one event in the sentence (\emph{``Later, they subsided."}). These relate an event in their sentence with the discourse's current temporal focus -- for example, document creation time, the previous sentence's main event, or reference time~\cite{reichenbach1947tenses,dowty1979word}. In a more complex case, such as Example~\ref{ex:later}, we suggest that two temporal links are present. First, \emph{Later} is attached to the current focus, as is \emph{surveyed}. Secondly, \emph{after} describes the relation between \emph{the storm} and \emph{surveyed}.

\begin{example}
\emph{It rained heavily. Later, \underline{after} the storm, we surveyed the damage.}
\label{ex:later}
\end{example}

Sometimes a signal may appear to only take one argument, when the other is (implicitly) reference time. For example, \emph{afterwards} and \emph{after that} are temporally equivalent, though \emph{afterwards} only takes one extra argument.

Signal surface forms have a compound structure consisting of a \textbf{head} and an optional \textbf{qualifier}. The head describes the temporal operation of the signal phrase and the qualifier modifies or clarifies this operation. An example of an unqualified signal expression is \emph{after}, which provides information about the nature of a temporal link, but does not say anything about the absolute or relative magnitude of the temporal separation of its arguments. We can elaborate on this with phrases which give qualitative information about the relative size of temporal separation between events (such as \emph{very shortly after}), or which give a specific separation between events using a duration as a modifying phrase (e.g. \emph{two weeks after}). 

\subsection{TimeML and TimeBank}

TimeML~\cite{pustejovsky2004specification} is a temporal annotation language. It may be used to annotate events, time expressions or {\it timex's}  (times, dates, durations), temporal relations between events and times (such as {\it before} or {\it during}), and signal expressions -- words or phrases (such as conjunctions, adverbials) that provide information about temporal relations. TimeBank ~\cite{pustejovsky2003timebank} is currently the largest TimeML-annotated gold standard corpus available, including over 6~000 temporal relation annotations, as well as events, times and signals. It consists of around 65~000 tokens of English newswire text.

TimeML offers the following definition of temporal signal. From the annotation guidelines\footnote{See http://timeml.org/site/publications/timeMLdocs /annguide\_1.2.1.pdf .}:

\begin{quote}
\textit{A signal is a textual element that makes explicit the relation holding between two entities (timex and event, timex and timex, or event and event). Signals are generally:
\begin{itemize}
\item Temporal prepositions: on, in, at, from, to, before, after, during, etc.
\item Temporal conjunctions: before, after, while, when, etc.
\item Prepositions signaling modality: to.
\item Special characters: ``-" and ``/", in temporal expressions denoting ranges.
\end{itemize}
}
\end{quote}

In cases where a specific duration occurs as part of a complex qualifier-head temporal signal, e.g. {\it two weeks after},  TimeBank has followed the convention that the signal head alone is annotated as a signal and the qualifier is annotated as a TIMEX of type \textsc{duration}.

\subsection{Previous Work}\label{previous}
Signals help create well-structured discourse. Temporal signals can
provide context shifts and orderings~\cite{hitzeman1997semantic}.
These signal expressions therefore work as discourse segmentation
markers~\cite{hodac2008temporal}. It has been shown that correctly
including such explicit markers make texts easier for human readers to
process~\cite{bestgen1999temporal}.

Some prior work has approached linguistic characterisation of signals.
Br\'ee et al.~\shortcite{bree1986temporal} performed a 
study of temporal conjunctions and prepositions and suggested rules for
discriminating temporal from non-temporal uses of signal expressions
that fall into these classes. However, this work is purely theoretical and
not a corpus-based study. Schl\"uter~\shortcite{schluter2001temporal} identifies signal
expressions used with the present perfect and compares their frequency
in British and US English. Vlach~\shortcite{vlach1993temporal}
presents a semantic framework that deals with duratives when used as
signal modifiers (see Section~\ref{signalstructure}). Br\'ee et
al.~\shortcite{bree1993towards} later describe the ambiguity of nine
temporal prepositions in terms of their roles as temporal signals. Our
work differs from the literature in that is it the first to be based
on gold standard annotations of temporal semantics and that it
encompasses all temporal signal expressions, not just those of a
particular grammatical class.

Intuitively, signal expressions contain temporal ordering
information that human readers can access easily. Once temporal
conjunctions are identified, existing semantic formalisms may be
applied to discourse semantics~\cite{dowty1979word}. It is however
ambiguous which temporal expression they attempt to
convey~\cite{hitzeman2005text}. Our work quantifies this ambiguity for
a subset of expressions.

Previous work applying temporal signals has been related to the
labeling of temporal links~\cite{min2007lcc} and question
answering~\cite{pustejovsky2005temporal,saquete2009enhancing}. In
particular, Lapata and Lascarides~\shortcite{lapata2006learning}
remove the temporal signal from sentences containing two temporally
connected clauses and attempt to learn sentence-level temporal
relations using the orderings suggested by the removed signal as
training data. Directly applying signals to the temporal relation
identification task, Derczynski and 
Gaizauskas~\shortcite{derczynski2010using} halved the error rate of
\texttt{TLINK} classification for \texttt{TLINK}s that have a signal
by adding features describing signals. This raised classification
accuracy from 62\% to 82\%.

\section{Signals in TimeBank}
\label{timebank}

In this section, we give a detailed profiling of temporal signals in the TimeBank corpus. Statistics are generated using the CAVaT~\cite{derczynski2010analysing} tool for TimeML-annotated corpus analysis.

First, we note that in TimeML  signals may be divided into three classes based on the type of relation they signal: temporal (tlink), sub-ordinating (slink) or aspectual (alink). The distribution of signals by class in Timebank is shown in Table~\ref{tab:v12sigusage}. For the rest of the paper we  discuss  temporal signals only.

\footnotesize
\begin{table}
\begin{center}
\begin{tabular}{ | l | c | }
\hline
Annotated \texttt{SIGNAL} elements & 758 \\
\hline
Signals used by a \texttt{TLINK} & 721 \\
Signals used by an \texttt{ALINK} & 1 \\
Signals used by a \texttt{SLINK} & 39 \\
\hline
\texttt{TLINK}s that use a \texttt{SIGNAL} & 787 \\
Signals used by more than one \texttt{TLINK} & 54 \\
\hline
\end{tabular}
\end{center}
\caption{How \texttt{<SIGNAL>} elements are used in TimeBank.}
\label{tab:v12sigusage}
\end{table}
\normalsize

\footnotesize
\begin{table}
\begin{center}
\begin{tabular}{ | l | r | }
\hline
\textbf{TLINKs per signal} & \textbf{Number of signals} \\
\hline
1 & 597 \\
2 & 41 \\
3 & 12 \\
5 & 1 \\
\hline
\end{tabular}
\caption{The number of TLINKs associated with each temporal signal word/phrase, in TimeBank. Signals not used on TLINKs (e.g. those used on aspectual or subordinate links, or for event cardinality) are excluded. The distribution is Zipfian.}
\label{tab:v12siglinks}
\end{center}
\end{table}
\normalsize

\footnotesize
\begin{table}
\begin{center}
\begin{tabular}{ | l | r | r |}
\hline
\textbf{Part of speech} & \textbf{Frequency} & \textbf {Proportion} \\
\hline
IN	&521	&77.3\%\\
RB	&73	&10.8\%\\
WRB	&53	&7.9\%\\
JJ	&14	&2.1\%\\
RBR	&5	&0.7\%\\
VBG	&4	&0.6\%\\
CC	&2	&0.3\%\\
RP	&1	&0.1\%\\
JJR	&1	&0.1\%\\
\hline
\end{tabular}
\caption{Distribution of part-of-speech in signals and the first word of multiword signals, using the Penn Treebank tag set.}
\label{tab:sig-pos}
\end{center}
\end{table}
\normalsize

\footnotesize
\begin{table*}
\begin{tabular}{| l | r | r | c | r | c |}
\hline
\textbf{Expression} & \textbf{Count in corpus} & \textbf{As signal} & \textbf{Proportion as signals} & \textbf{After curation} & \textbf{Proportion} \\
\hline 
in	& 1214	& 161	& 13.3\%	& 	&  \\
after	& 72	& 56	& 77.8\%	& 66	& 91.7\% \\
for	& 621	& 52	& 8.4\%	& 	&  \\
if	& 65	& 37	& 56.9\%	& 	&  \\
when	& 62	& 35	& 56.5\%	& 56	& 90.3\% \\
on	& 344	& 33	& 9.6\%	& 	&  \\
until	& 36	& 25	& 69.4\%	& 36	& 100.0\% \\
before	& 33	& 23	& 69.7\%	& 30	& 90.9\% \\
by	& 356	& 20	& 5.6\%	& 	&  \\
from	& 366	& 19	& 5.2\%	& 	&  \\
since	& 31	& 17	& 54.8\%	& 18	& 58.1\% \\
through	& 69	& 15	& 21.7\%	& 	&  \\
as	& 271	& 14	& 5.2\%	& 	&  \\
over	& 59	& 14	& 23.7\%	& 	&  \\
already	& 32	& 13	& 40.6\%	& 13	& 40.6\% \\
ended	& 21	& 13	& 61.9\%	& 	&  \\
during	& 19	& 13	& 68.4\%	& 	&  \\
at	& 311	& 11	& 3.5\%	& 	&  \\
previously	& 19	& 11	& 57.9\%	& 16	& 84.2\% \\
within	& 23	& 8	& 34.8\%	& 	&  \\
s	& 10	& 8	& 80.0\%	& 	&  \\
later	& 15	& 7	& 46.7\%	& 	&  \\
earlier	& 50	& 6	& 12.0\%	& 	&  \\
while	& 39	& 6	& 15.4\%	& 9	& 23.1\% \\
then	& 23	& 5	& 21.7\%	& 	&  \\
once	& 15	& 5	& 33.3\%	& 	&  \\
still	& 35	& 4	& 11.4\%	& 	&  \\
following	& 15	& 4	& 26.7\%	& 	&  \\
meanwhile	& 14	& 4	& 28.6\%	& 9	& 64.3\% \\
at the same time	& 6	& 4	& 66.7\%	& 	&  \\
to	& 1600	& 3	& 0.2\%	& 	&  \\
into	& 63	& 3	& 4.8\%	& 	&  \\
follows	& 4	& 3	& 75.0\%	& 	&  \\
subsequently	& 3	& 3	& 100.0\%	& 	&  \\
followed	& 10	& 2	& 20.0\%	& 4	& 40.0\% \\
former	& 16	& 0	& 0.0\%	& 12	& 75.0\% \\
\hline
\end{tabular}
\caption{Frequency of candidate signal expressions in TimeBank. We include counts of how often these occur as signal expressions both before and after manual curation.}
\label{tab:signal-use}
\end{table*}
\normalsize

\footnotesize
\begin{table*}
\begin{center}
\begin{tabular}{| l | c | r | r | r | r | r | r | r | r | r | r | r | r | r | r |}
\hline
\textbf{Signal Expression}	&  \textbf{TLINK count}	&  \begin{sideways}\textsc{after}\end{sideways}	&  \begin{sideways}\textsc{before}\end{sideways}	&  \begin{sideways}\textsc{begins}\end{sideways}	&  \begin{sideways}\textsc{begun\_by}\end{sideways}	&  \begin{sideways}\textsc{during}\end{sideways}	& \begin{sideways}\textsc{ended\_by}\end{sideways}	&  \begin{sideways}\textsc{ends}\end{sideways}	&  \begin{sideways}\textsc{iafter}\end{sideways}	&  \begin{sideways}\textsc{ibefore}\end{sideways}	&  \begin{sideways}\textsc{includes}\end{sideways}	&  \begin{sideways}\textsc{is\_included}\end{sideways}	&  \begin{sideways}\textsc{simultaneous}\end{sideways} \\
\hline
after	&  76	&  62	&  3	&  4	&  	&  	&  	&  5	&  2	&  	&  	&  	&   \\
when	&  57	&  16	&  3	&  1	&  2	&  1	&  	&  	&  1	&  1	&  9	&  9	&  14 \\
until	&  37	&  4	&  7	&  1	&  	&  	&  21	&  1	&  1	&  2	&  	&  	&   \\
before	&  36	&  1	&  28	&  2	&  	&  	&  1	&  2	&  1	&  1	&  	&  	&   \\
since	&  19	&  9	&  1	&  2	&  7	&  	&  	&  	&  	&  	&  	&  	&   \\
already	&  13	&  	&  6	&  	&  	&  	&  	&  	&  	&  	&  4	&  3	&   \\
previously	&  18	&  6	&  12	&  	&  	&  	&  	&  	&  	&  	&  	&  	&   \\
while	&  9	&  	&  	&  	&  	&  	&  	&  	&  	&  	&  	&  	&  9 \\
meanwhile	&  9	&  	&  1	&  	&  	&  2	&  	&  	&  	&  	&  	&  1	&  5 \\
followed	&  4	&  2	&  2	&  	&  	&  	&  	&  	&  	&  	&  	&  	&   \\
former	&  12	&  	&  12	&  	&  	&  	&  	&  	&  	&  	&  	&  	&   \\
\hline
\end{tabular}
\caption{Signal expressions and the TimeML relations that they can denote, ordered as per Table~4 for comparison. Counts do not match because a single signal expression can support more than one temporal link.}
\label{tab:signalusage}
\end{center}
\end{table*}
\normalsize

\subsection{Additional Annotation}
\label{reannotation}
Upon examination of the non-annotated instances of words that often occur as a temporal signal (such as \emph{after}) it became evident that TimeBank's signals are under-annotated. As we are certain of some annotation errors in the source data, we revisited the original annotations. A subset of signal words was selected for re-annotation. This set consisted of signals that were ambiguous (occurred temporally close to 50\% of the time) or that we expected contained, based on informal observations, would yield a number of missed temporal annotations. All temporal instances of these words were re-annotated with TimeML, adding \texttt{EVENT}s, \texttt{TIMEX3}s and \texttt{TLINK}s where necessary to create a signalled \texttt{TLINK}.

A single annotator checked the source documents and annotated 70 extra signals, as well as adding 34 events, 1 temporal expression and 49 extra temporal links. 

\subsection{Proportion of Temporal Relations with Signals}

TimeBank contains 6~418 TLINKs (6~467 after re-annotation) of which 718 (787) are explicitly indicated by a temporal signal -- 11.2\% (12.2\%). This provides an answer to the first question we posed in Section 1. Thus while ability to successfully detect temporal signals will not solve the problem of assigning temporal relations, it is likely to make a noticeable difference (see Derczynski and Gaizauskas~\shortcite{derczynski2010using}). Perhaps of more interest is that so few temporal relations are explicitly signalled -- we must look elsewhere for an explanations of how temporal relations are conveyed in natural language.

While many TLINKs do not have any associated temporal signal it is also the case that some temporal signals are associated with more than one TLINK. Table~\ref{tab:v12siglinks} shows details of just how signals are being used by TLINKs. 

\subsection{Temporal vs Non-temporal Uses}

The semantic function that a temporal signal expression performs is that of relating two temporal  entities. However, the words that can function as temporal signals also play other roles.  


Table~\ref{tab:signal-use} details the distribution of  expressions that are found as temporal signals more than twice (after re-annotation)  in TimeBank. 
The most frequent signal word was ``in", accounting for 24.8\% of all signal-using \texttt{TLINK}s. However, only 13.3\% of occurrences of the word ``in" have a temporal sense. The word ``after" is far more likely (91.7\% of all occurrences) to have a temporal sense. 
In total TimeBank contains 62 unique signal words and phrases (ignoring case) and
of these, over half (36) are also found in Table~\ref{tab:signal-use} . 

As an aside, note that any thought that temporal signals might be easily picked out based on word class may be dispelled by examining the distribution of parts-of-speech possessed by temporal signals -- see Table~\ref{tab:sig-pos}.



\subsection{Relation Ambiguity}

The nature of the temporal relation described by a signal is not constant, though each signal tends to describe a particular relation type most often. Table~\ref{tab:signalusage} gives an excerpt of data showing which temporal relations are made explicit by each signal expression. The variation in relation type associated with a signal is not as great as it might appear as the assignment of temporal relation type has an element of arbitrariness --  one may choose to annotate a \textsc{before} or \textsc{after} relation for the same event pair by simply reversing the temporal link's argument order, for example. Nevertheless, it is possible to draw  useful information from the table; for example, one can see that \emph{meanwhile} is much more likely to suggest some sort of temporal overlap between events than an ordering where arguments occur discretely.

\section{Per-expression details}

We chose to curate signal annotations in TimeBank for a subset of candidate signal expressions (as described in Section~\ref{reannotation}). During this curation, we attempted to determine distinguishing features that could aid automatic discrimination of temporal from non-temporal sense of the expressions. Details of our findings are given below.

\paragraph{Previously}

TimeBank contains eight instances of the word that were not annotated as a signal. Of these, all were being used as temporal signals. The word only takes one event or time as its direct argument, which is placed temporally before an event or time that is in focus. For example:

\emph{``X reported a third-quarter loss, citing a previously announced capital restructuring program"}

In this sentence, the second argument of \emph{previously} is \emph{``announced"}, which is temporally situated before its first argument (\emph{``reported"}). When \emph{previously} occurs at the top of a section, the temporal element that has focus is either document creation time or, if one has been specified in previous discourse, the time currently in focus.

\paragraph{After}

Of the nineteen instances of this word not annotated as temporal, only three were actually non-temporal. The cases that were non-temporal were a different sense of the word. The temporal signals are adverbial, with a temporal function. Two non-temporal cases used a positional sense. The last case was in a phrasal verb \emph{to go after}; \emph{``whether we would go after attorney's fees"}.

%

%

\paragraph{When}

There are 35 annotated and 27 non-annotated occurrences of this phrase. It indicates either an overlap between intervals, or a point relation that matches an interval's start. Twenty-three of the twenty-seven non-annotated occurrences are used as temporal signals. Two of the remaining four are in negated phrases and not used to link an interval pair. for example, \emph{``did not say \underline{when} the reported attempt occurred"}. The other two are used in context setting phrases, e.g. \emph{``we think he is someone who is capable of rational judgements \underline{when} it comes to power"}, which are not temporal in nature.

\paragraph{While}

The cases of \emph{while} that have not been annotated as a signal -- the majority class, 33 to 6 -- are often used in a contrastive sense. This does suggest that the connected events have some overlap, often between statives. For example, \emph{``But \underline{while} the two Slavic neighbours see themselves as natural partners, their relations since the breakup of the Soviet Union have been bedeviled"}. As two states described in the same sentences are likely to temporally overlap and any events or times outside or bounding these states will be related to the state, it is unlikely that any contribution to \texttt{TLINK} annotation would be made by linking the two states with a ``roughly simultaneous" relation; the closest suitable label is TempEval's \textsc{overlap} relation~\cite{verhagen2010semeval}.

\begin{example}
\emph{``nor can the government easily back down on promised protection for a privatized company \underline{while} it proceeds with \ldots"}
\label{ex:incorrect-ftag}
\end{example}

The cases of \emph{while} that were not of this sense were easier to annotate. Sometimes it was used as a temporal expression; \emph{``for a while"}. Other times, it was not used in a contrastive sense, but instead as irrealis -- see Example~\ref{ex:incorrect-ftag}. The four cases of non-contrastive usage were annotated as temporal signals.

\begin{figure}
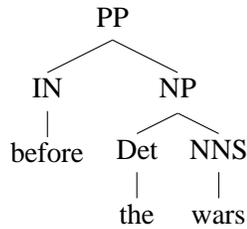

\Tree [.PP [.IN before ] [.NP [.Det the ] [.NNS wars ] ] ]
\caption{An example of the common syntactic surroundings of a \emph{before} signal.}
\label{fig:parsetree-before}
\end{figure}

\paragraph{Before}

Three of the ten negative examples are correctly annotated. They are \emph{before} in the spatial sense of ``in front of" (as in \emph{``The procedures are to go \underline{before} the Security Council next week"}) and also a logical before that does not link instantiated or specific events (\emph{``\underline{before} taxes"}). The remaining seven unannotated examples of the word are all temporal signals. These directly precede either an NP describing a nominalised event, or directly precede a subordinate clause (e.g. \texttt{[IN before, S]} -- see Figure~\ref{fig:parsetree-before}).

\paragraph{Until}

All eleven non-annotated instances of \emph{until} should have been annotated as temporal signals. This word suggests a TimeML \textsc{ibefore} relation, unless qualified otherwise by something like ``not until" or ``at least until".


\paragraph{Already}

There were thirteen positive examples of \emph{already}. All of the non-annotated examples had a non-temporal sense as per our description of temporal signals. The word tends to be used for emphasis, but can also suggest a broad ``\textsc{before} DCT" position, which goes without saying for any past and present tensed events. As \emph{already} can be removed without changing the temporal links present in a sentence, we have not annotated any more examples of this beyond the thirteen present in TimeBank.


\paragraph{Meanwhile}

This word tends to refer to a reference or event time introduced earlier in discourse, often from the same sentence. As well as a temporal sense, it can have a contrastive ``despite"-like meaning. \emph{Meanwhile} tends to refer more to previous actions, instead of states specified in immediately prior sentences. Sometimes \emph{meanwhile} is used with no previous temporal reference. In these cases, the implicit argument is DCT. Five of the ten non-annotated \emph{meanwhile}s were temporal signals.


\paragraph{Again}

This word shows recurrence and is always used for this purpose where it occurs in TimeBank not annotated as a temporal signal. No instances of \emph{``again"} were annotated.

\begin{figure}
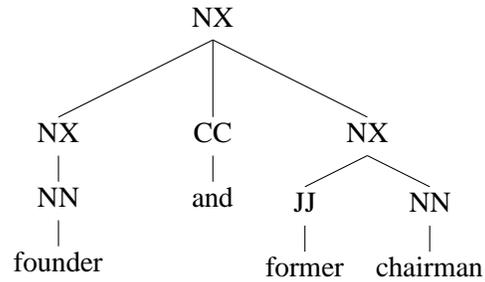

\Tree [.NX
        [.NX [.NN founder ] ]
        [.CC and ]
        [.NX [.JJ former ] [.NN chairman ] ] ]
\caption{Example of a non-annotated signal (\emph{former}) from TimeBank's \texttt{wsj\_0778.tml.}}
\label{fig:parsetree-former}
\end{figure}

\paragraph{Former}

This word indicates a state that persisted before DCT or current speech time and has now finished. Generally the construction that is found is an NP, which contains an optional determiner, followed by \emph{former} and then a substituent NP which may be annotated as an \texttt{EVENT} of class \textsc{state}. This configuration suggests a \texttt{TLINK} that places the event \textsc{before} the state's utterance. 

\begin{example}
\emph{``The San Francisco sewage plant was named in honour of former President Bush."}
\label{ex:sewage}
\end{example}

In Example~\ref{ex:sewage}, there is a \textsc{state}-class event -- \emph{President} -- that at one time has applied to the named entity \emph{Bush}. The signal expression \emph{former} indicates that this state terminated \textsc{before} the time of the sentence's utterance.

Three-quarters of the non-annotated instances of \emph{former} in TimeBank are temporal signals.

\paragraph{Recently}

Although \emph{recently} is a temporal adverb, it can only be to applied simple or anterior tensed verbs (using Reichenbach's tense nomenclature). In the corpus, these are only seen in reported speech or of verbal events that happened before DCT. \emph{Recently} adds a qualitative distance between event and utterance time, but is of reduced use when we can already use tense information. The phrase ``Until recently" appears awkward when cast as a temporal signal but can be interpreted as ``\textsc{before} DCT", with the interval's endpoint being close to DCT. In this case, recently functions as a temporal expression, not a signal. Only one of the non-annotated \emph{recently}s in TimeBank is a temporal signals. The exception, \emph{``More recently"}, includes a comparative and is annotated as a \texttt{TIMEX3}; both this phrase and, e.g., \emph{``less recently"} suggest a relation to a previously-mentioned (and in-focus) past event. As a result, we posit that \emph{recently} behaves as an abstract temporal point (as seen in the behaviour of \emph{``until recently"}). Structures such as \emph{[comparative] recently} may be interpreted as a qualified temporal signal, as they convey information about the relative ordering of the event that they dominate vent compared with a previously mentioned interval.

\section{Conclusion}

We have provided a characterisation of temporal signal expressions. In an analysis of the TimeBank corpus we have shown what proportion of temporal relations are explicitly signalled by these expressions and have given quantitative descriptions of how ambiguous these phrases are, both regarding  their temporal/non-temporal senses and  the type of temporal relation that they convey.

\subsection{Acknowledgments}
The first author would like to acknowledge the UK Engineering and Physical Science Research Council's support in the form of a doctoral studentship.

\bibliographystyle{lrec2006}
\bibliography{sigcorp}

\end{document}